%% file: 0main.tex
\documentclass[11pt,a4paper]{article}
\usepackage{times,latexsym}
\usepackage{url}
\usepackage[T1]{fontenc}

\usepackage[acceptedWithA]{tacl2018v2}

\usepackage{xspace,mfirstuc,tabulary}

\newif\iftaclinstructions
\taclinstructionsfalse 
\iftaclinstructions
\renewcommand{\confidential}{}
\renewcommand{\anonsubtext}{(No author info supplied here, for consistency with
TACL-submission anonymization requirements)}
\newcommand{\instr}
\fi

\iftaclpubformat 

\else

\fi

\usepackage{amsmath,amsfonts,amssymb}
\usepackage{booktabs}
\usepackage{algorithm}
\usepackage{algorithmic}
\usepackage{xspace}
\usepackage{multirow}
\usepackage{verbatim}
\usepackage{pifont}
\newcommand{\cmark}{\ding{51}}%
\usepackage{subfig}
\usepackage{tcolorbox}
\usepackage[export]{adjustbox}
\usepackage{xcolor}
\usepackage{enumitem}
\usepackage{bm}
\usepackage{tikz,pgfplots}
\usepackage{tcolorbox}
\usepackage{caption}
\usepackage{graphicx}
\usepackage{booktabs}
\usepackage{xcolor,pifont}

\definecolor{green}{rgb}{0.0, 0.65, 0.31}
\newcommand*\colourcheck[1]{%
  \expandafter\newcommand\csname #1check\endcsname{\textcolor{#1}{\ding{51}}}%
}
\colourcheck{green}

\usepackage{array}
\newcolumntype{?}{!{\vrule width 1pt}}

\usetikzlibrary{arrows,decorations.markings,patterns}
\usetikzlibrary{arrows.meta}

\definecolor{one}{HTML}{222222}
\definecolor{two}{HTML}{56c1ff}
\definecolor{three}{HTML}{ff968d}
\definecolor{four}{HTML}{bdcc2f}
\definecolor{five}{HTML}{adadad}
\definecolor{six}{HTML}{FB8B24}
\definecolor{junglegreen}{rgb}{0.16, 0.67, 0.53}

\newcommand{\DRAIL}{\textsc{DRaiL}\xspace}
\newcommand{\relnets}{\textsc{RelNets}\xspace}
\newcommand{\glrelnets}{\textsc{RelNets}\xspace}

\newcommand{\Global}{\textsc{Global}\xspace}
\newcommand{\Inference}{\textsc{JointInf}\xspace}
\newcommand{\Independent}{\textsc{IndNets}\xspace}
\newcommand{\Endtoend}{\textsc{E2E}\xspace}

\newcommand{\PRED}[1]{$\mathtt{#1}$}
\newcommand{\NPRED}[1]{$\neg\mathtt{#1}$}

\def\argmax{\operatornamewithlimits{arg\,max}}

\DeclareMathSymbol{\mathdblquotechar}{\mathalpha}{letters}{`"}

\newcommand{\mathdblquote}{\mathtt{\mathdblquotechar}}
\begingroup\lccode`~=`"\lowercase{\endgroup
  \let~\mathdblquote
}
\AtBeginDocument{\mathcode`"="8000 }

\title{Modeling Content and Context with Deep Relational Learning}

\author{Maria Leonor Pacheco \and Dan Goldwasser \\
         Department of Computer Science \\
         Purdue University \\ West Lafayette, IN 47907 \\
         \texttt{\{pachecog, dgoldwas\}@purdue.edu}}

\date{}

\begin{document}
\maketitle
\begin{abstract}
Building models for realistic natural language tasks requires dealing with long texts and accounting for complicated structural dependencies. Neural-symbolic representations have emerged as a way to combine the reasoning capabilities of symbolic methods, with the expressiveness of neural networks. However, most of the existing frameworks for combining neural and symbolic representations have been designed for classic relational learning tasks that work over a universe of symbolic entities and relations. In this paper, we present \DRAIL, an open-source declarative framework for specifying deep relational models, designed to support a variety of NLP scenarios. Our framework supports easy integration with expressive language encoders, and provides an interface to study the interactions between representation, inference and learning. 
\end{abstract}

\input{1intro.tex}
\input{2related.tex}

\input{3framework.tex}
\input{4learning.tex}
\input{7experiments.tex}
\input{9conclusion.tex}

\input{10acknowledgments.tex}

\bibliography{references}
\bibliographystyle{acl_natbib}

\newpage
\input{appendix}

\end{document}

%% file: 1intro.tex

\section{Introduction}
   

Understanding natural language interactions in realistic settings requires models that can deal with noisy textual inputs, reason about the dependencies between different textual elements and leverage the dependencies between textual content and the context from which it emerges. Work in linguistics and anthropology has defined context as a frame that surrounds a focal communicative event and provides resources for its interpretation \cite{contextualization-92,Duranti1992}. 

\begin{figure}%
    \centering
        \includegraphics[width=0.5\textwidth]{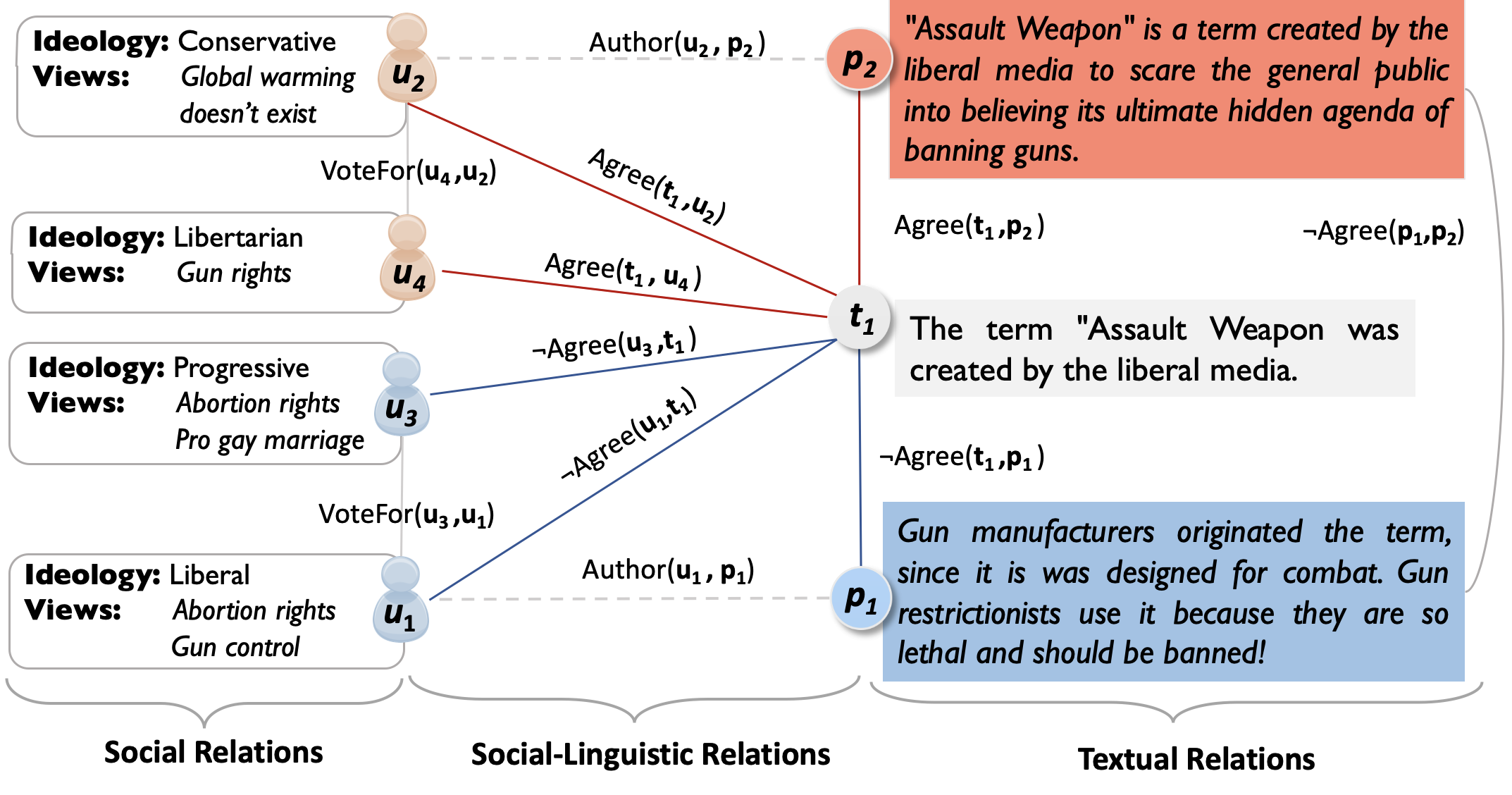}
        \captionof{figure}{Example debate}
        \label{fig:exampleDebate}
\end{figure}

As a motivating example, consider the interactions in the debate network  described in Fig.~\ref{fig:exampleDebate}. Given a debate claim ($t_1$), and two consecutive posts debating it ($p_1, p_2$), we define a textual inference task, determining whether a pair of text elements hold the same stance in the debate (denoted using the relation \PRED{Agree(X,Y)}). This task is similar to other textual inference tasks~\cite{bowman2015large} which have been successfully approached using complex neural representations~\cite{peters2018deep,devlin2018bert}. In addition, we can leverage the dependencies between these decisions.  For example, assuming that one post agrees with the debate claim ({\footnotesize\PRED{Agree(t_1,p_2)}}), and the other one does not ({\footnotesize\NPRED{Agree(t_1,p_1)}}), the disagreement between the two posts can be inferred:  {\small \PRED{\neg Agree(t_1,p_1)\wedge Agree(t_1,p_2) \rightarrow \neg Agree(p_1,p_2)}}. Finally, we consider the \textit{social context} of the text. The disagreement between the posts can reflect a difference in the perspectives their authors hold on the issue. While this information might not be directly observed, it can be inferred using the authors' social interactions and behavior.
Given the principle of social homophily~\cite{mcpherson2001birds}, stating that people with strong social ties are likely to hold similar views and authors' perspectives can be captured by representing their social interactions. Exploiting this information requires models that can align the social representation with the linguistic one.

Motivated by these challenges, we introduce \DRAIL\footnote{\url{https://gitlab.com/purdueNlp/DRaiL/}}, a Deep Relational Learning framework, which uses a combined neuro-symbolic representation for modeling the interaction between multiple decisions in relational domains. Similar to other neuro-symbolic approaches~\cite{mao2019neuro,DBLP:journals/jair/CohenYM20} our goal is to exploit the complementary strengths of the two modeling paradigms. Symbolic representations, used by logic-based systems and by probabilistic graphical models~\cite{richardson2006markov,bac:jmlr17}, are interpretable, and allow domain experts to directly inject knowledge and constrain the learning problem. Neural models capture dependencies using the network architecture and are better equipped to deal with noisy data, such as text. However, they are often difficult to interpret and constrain according to domain knowledge.

Our main design goal in \DRAIL is to provide a generalized tool, specifically designed for NLP tasks. Existing approaches designed for classic relational learning tasks~\cite{DBLP:journals/jair/CohenYM20}, such as knowledge graph completion, are not equipped to deal with the complex linguistic input. While others are designed for very specific NLP settings such as word-based quantitative reasoning problems~\cite{NIPS2018_7632} or aligning images with text~\cite{mao2019neuro}. We discuss the differences between \DRAIL and these approaches in Section~\ref{sec:related}. 
While the examples in this paper focus on modelings various argumentation mining tasks and their social and political context, the same principles can be applied to wide array of NLP tasks with different contextualizing information, such as images that appear next to the text, or prosody when analyzing transcribed speech, to name a few examples.

 
%
\DRAIL uses a declarative language for defining deep relational models. Similar to other declarative languages~\cite{richardson2006markov,bac:jmlr17}, it allows users to inject their knowledge by specifying dependencies between decisions using first-order logic rules, which are later compiled into a factor graph with neural potentials.  
In addition to probabilistic inference, \DRAIL also models dependencies using a \textit{distributed knowledge representation}, denoted \relnets, which provides a shared representation space for entities and their relations, trained using a relational multi-task learning approach. This provides a mechanism for explaining symbols, and aligning representations from different modalities. 
Following our running example, ideological standpoints, such as \PRED{Liberal} or \PRED{Conservative}, are discrete entities embedded in the same space as textual entities and social entities. These entities are initially associated with users, however using \relnets this information will propagate to texts reflecting these ideologies, by exploiting the relations that bridge social and linguistic information (see Fig.~\ref{fig:exampleDebate}). 

 

To demonstrate \DRAIL's modeling approach, we introduce the task of \textit{open-domain stance prediction with social context}, which combines social networks analysis and textual inference over complex opinionated texts, as shown in Fig. \ref{fig:exampleDebate}.
%
We complement our evaluation of \DRAIL with two additional tasks, issue-specific stance prediction, where we identify the views expressed in debate forums with respect to a set of fixed issues~\cite{Walker:2012:SCU:2382029.2382124}, and argumentation mining~\cite{Stab2017PAS}, a document-level discourse analysis task. 
 




 

%% file: 2related.tex
\section{Related Work}\label{sec:related}

\begin{table*}[t]
    \centering
    \resizebox{\textwidth}{!}{%
    \begin{tabular}{l?c|c|c|c|c?c|c|c|c|c?c}
    \toprule
   \multirow{3}{*}{\textbf{System}} & \multicolumn{5}{c?}{\textbf{Symbolic Features}} & \multicolumn{5}{c?}{\textbf{Neural Features}} &  \\ \cline{2-11}
      & Symbolic & Raw & Decla- &  Prob/Logic & Rule & Embed. &  End-to-end & Backprop. to & Architecture  &  Multi-Task & \textbf{Open} \\
     & Inputs & Inputs & rative & Inference & Induction & Symbols & Neural  &  Encoders &  Agnostic & Learning & \textbf{Source} \\
     \hline
     
     MLN & \cmark & & \cmark & \cmark &  & &  & & & & \cmark \\
     FACTORIE & \cmark & & & \cmark &  & &  & & & & \cmark \\ 
     CCM & \cmark & \cmark & \cmark & \cmark &  & &  & & & & \cmark \\
     PSL & \cmark & & \cmark & \cmark &  & &  & & & & \cmark \\ 
     \hline 
     
     LRNNs & \cmark & & & & & \cmark &  \cmark  &  &  &  & \\
     RelNNs & \cmark & & & & & \cmark & \cmark  &  & & & \cmark \\ 
     \hline 
     
     LTNs & \cmark & &  \cmark & \cmark & & \cmark & \cmark &  & & & \cmark \\
     TensorLog & \cmark & & \cmark & \cmark & & \cmark & \cmark &  & & &  \cmark \\ 
     \hline 
     
     NTPs & \cmark & & \cmark & \cmark & \cmark & \cmark & \cmark &  & & &  \cmark \\
     Neural LP & \cmark & &  &  & \cmark & \cmark & \cmark &  & & & \cmark \\
     DRUM & \cmark & &  &  & \cmark & \cmark & \cmark &  & & &  \cmark \\
     NLMs & \cmark & &  &  & \cmark & \cmark & \cmark &  & & &  \cmark \\ 
     \hline
     DeepProbLog & \cmark & \cmark & \cmark & \cmark & \cmark & \cmark &  & \cmark & \cmark & & \cmark \\ 
     DPL & \cmark & \cmark & \cmark & \cmark & & \cmark & & \cmark & \cmark & & \cmark \\
     DLMs & \cmark & \cmark & \cmark & \cmark & & \cmark & \cmark & \cmark & \cmark & & \\ 
     {\bf \DRAIL} & \cmark & \cmark & \cmark & \cmark & & \cmark & & \cmark & \cmark & \cmark  & \cmark\\
    \bottomrule
    \end{tabular}}
    \caption{Comparing Systems}\label{tab:feature_matrix}
\end{table*}

In this section, we survey several lines of work dealing with symbolic, neural and hybrid representations for relational learning. 

\subsection{Languages for Graphical Models}


Several high level languages for specifying graphical models have been suggested. BLOG~\cite{milch2005blog} and CHURCH~\cite{goodman2012church}  were suggested for generative models. For discriminative models, we have Markov Logic Networks (MLNs)~\cite{richardson2006markov} and Probabilistic Soft Logic (PSL)~\cite{bac:jmlr17}. Both PSL and MLNs combine logic and probabilistic graphical models in a single representation, where each formula is associated with a weight, and the probability distribution over possible assignments is derived from the weights of the formulas that are satisfied by such assignments. Like DRaiL, PSL uses formulas in clausal form (specifically collections of horn clauses). The main difference between \DRAIL and these languages is that, in addition to graphical models, it uses distributed knowledge representations to represent dependencies. Other discriminative methods include  FACTORIE~\cite{mccallum09:factorie:}, an imperative language to define factor graphs, Constraints Conditional Models (CCMs) ~\cite{RizzoloRo10,kordjamshidisaul} an interface to enhance linear classifiers with declarative constraints, and ProPPR~\cite{DBLP:conf/cikm/WangMC13} a probabilistic logic for large databases that approximates local groundings using a variant of personalized PageRank.

\subsection{Node Embedding and Graph Neural Nets}\label{sec:related_node_embedding}
A recent alternative to graphical models is to use neural nets to  represent and learn over relational data, represented as a graph. Similar to \DRAIL's \relnets, the learned node representation can be trained by several different prediction tasks. However, unlike \DRAIL, these methods do not use probabilistic inference to ensure consistency.

Node embeddings approaches~\cite{Perozzi:2014:DOL:2623330.2623732,tang2015line,tridnr,node2vec-kdd2016,tu-etal-2017-cane} learn a feature representation for nodes capturing graph adjacency information, such that the similarity in the embedding space of any two nodes is  proportional to their graph distance and overlap in neighbouring nodes. Some frameworks~\cite{tridnr,xiao2017ssp,tu-etal-2017-cane} allow nodes to have textual properties, which provide an initial feature representation when learning to represent the graph relations. When dealing with multi-relational data, such as knowledge graphs, both the nodes and the edge types are embedded ~\cite{transE,wang2014knowledge,pmlr-v48-trouillon16,sun2018rotate}. Finally, these methods learn to represent nodes and relations based on pair-wise node relations, without representing the broader graph context in which they appear. Graph neural nets~\cite{kipf2016semi,hamilton2017inductive,velivckovic2017graph} create contextualized node representations by recursively aggregating neighbouring nodes' information.

\subsection{Hybrid Neural-Symbolic Approaches}
Several recent systems explore ways to combine neural and symbolic representations in a unified way. We group them into five categories.
 
\textbf{Lifted rules to specify compositional nets}. These systems use an end-to-end approach and learn relational dependencies in a latent space. Lifted Relational Neural Networks (LRNNs) \cite{DBLP:journals/jair/SourekAZSK18} and RelNNs \cite{kazemi18} are two examples. These systems map observed ground atoms, facts and rules to specific neurons in a network and define composition functions directly over them. While they provide for a modular abstraction of the relational inputs, they assume all inputs are symbolic and do not leverage expressive encoders. 

\textbf{Differentiable inference}. These systems identify classes of logical queries that can be compiled into differentiable functions in a neural network infrastructure. In this space we have Tensor Logic Networks (TLNs) \cite{ijcai2017-221} and TensorLog \cite{DBLP:journals/jair/CohenYM20}. Symbols are represented as row vectors in a parameter matrix. The focus is on implementing reasoning using a series of numeric functions. 

\textbf{Rule induction from data}.
These systems are designed for inducing rules from symbolic knowledge bases, which is not in the scope of our framework. In this space we find Neural Theorem Provers (NTPs) \cite{NIPS2017_6969}, Neural Logic Programming \cite{NIPS2017_6826}, DRUM \cite{NIPS2019_9669} and Neural Logic Machines (NLMs) \cite{48065}. NTPs use a declarative interface to specify rules that add inductive bias and perform soft proofs. The other approaches work directly over the database. 

\textbf{Deep classifiers and probabilistic inference}. These systems propose ways to integrate probabilistic inference and neural networks for diverse learning scenarios. DeepProbLog \cite{NIPS2018_7632} extends the probabilistic logic programming language ProbLog to handle neural predicates. They are able to learn probabilities for atomic expressions using neural networks. The input data consists of a combination of feature vectors for the neural predicates, together with other probabilistic facts
and clauses in the logic program. Targets are only given at the output side of the probabilistic reasoner, allowing them to learn each example with respect to a single query.  
On the other hand, Deep Probabilistic Logic (DPL) \cite{wang-poon-2018-deep} combines neural networks with probabilistic logic for indirect supervision. They learn classifiers using neural networks and use probabilistic logic to introduce distant supervision and labeling functions. Each rule is regarded as a latent variable, and the logic defines a joint probability distribution over all labeling decisions. Then, the rule weights and the network parameters are learned jointly using variational EM. In contrast, \DRAIL focuses on learning multiple interdependent decisions from data, handling and requiring supervision for all unknown atoms in a given example. 
%
Lastly, Deep Logic Models (DLMs) \cite{marra19dlm} learn a set of parameters to encode atoms in a probabilistic logic program. Similarly to \citealt{ijcai2017-221} and \citealt{DBLP:journals/jair/CohenYM20}, they use differentiable inference, allowing the model to be trained end-to-end. Like \DRAIL, DLMs can work with diverse neural architectures and backpropagate back to the base classifiers. The main difference between DLMs and \DRAIL is that \DRAIL ensures representation consistency of entities and relations across all learning tasks by employing \relnets. 

\textbf{Deep structured models}. More generally, deep structured prediction approaches have been successfully applied to various NLP tasks such as NER and dependency parsing~\cite{chen2014fast,weiss-EtAl:2015:ACL-IJCNLP,ma2016end,lample2016neural,kiperwasser2016simple,malaviya2018neural}. When the need arises to go beyond sentence-level, some works combine the output scores of independently trained classifiers using inference~\cite{ beltagy-etal-2014-probabilistic,sridhar2015joint,liu-etal-2016-leveraging,subramanian-etal-2017-joint,ning2018joint}, while others implement joint learning for their specific domains \cite{niculae-etal-2017-argument,han-etal-2019-joint}. Our main differentiating factor is that we provide a general interface that leverages FOL clauses to specify factor graphs and express constraints. 


To summarize these differences, we outline a feature matrix in Tab. \ref{tab:feature_matrix}. Given our focus in NLP tasks, we require a neural-symbolic system that (1) allows us to integrate state-of-the-art text encoders and NLP tools, (2) supports structured prediction across long texts, (3) lets us combine several modalities and their representations (e.g. social and textual information) and (4) results in an explainable model where domain constraints can be easily introduced.

%% file: 3framework.tex
\section{The \DRAIL Framework}


%
%


\DRAIL was designed for supporting complex NLP tasks. Problems can be broken down into domain-specific atomic components (which could be words, sentences, paragraphs or full documents, depending on the task), and dependencies between them, their properties and contextualizing information about them can be explicitly modeled. 

%
In \DRAIL dependencies can be modeled over the predicted output variables (similar to other probabilistic graphical models), as well as over the neural representation of the atoms and their relationships in a shared embedding space.
This section explains the framework in detail. We begin with a high-level overview of \DRAIL and the process of moving from a declarative definition to a predictive model. 

\begin{figure}[t]%
    \centering
    \includegraphics[width=0.5\textwidth]{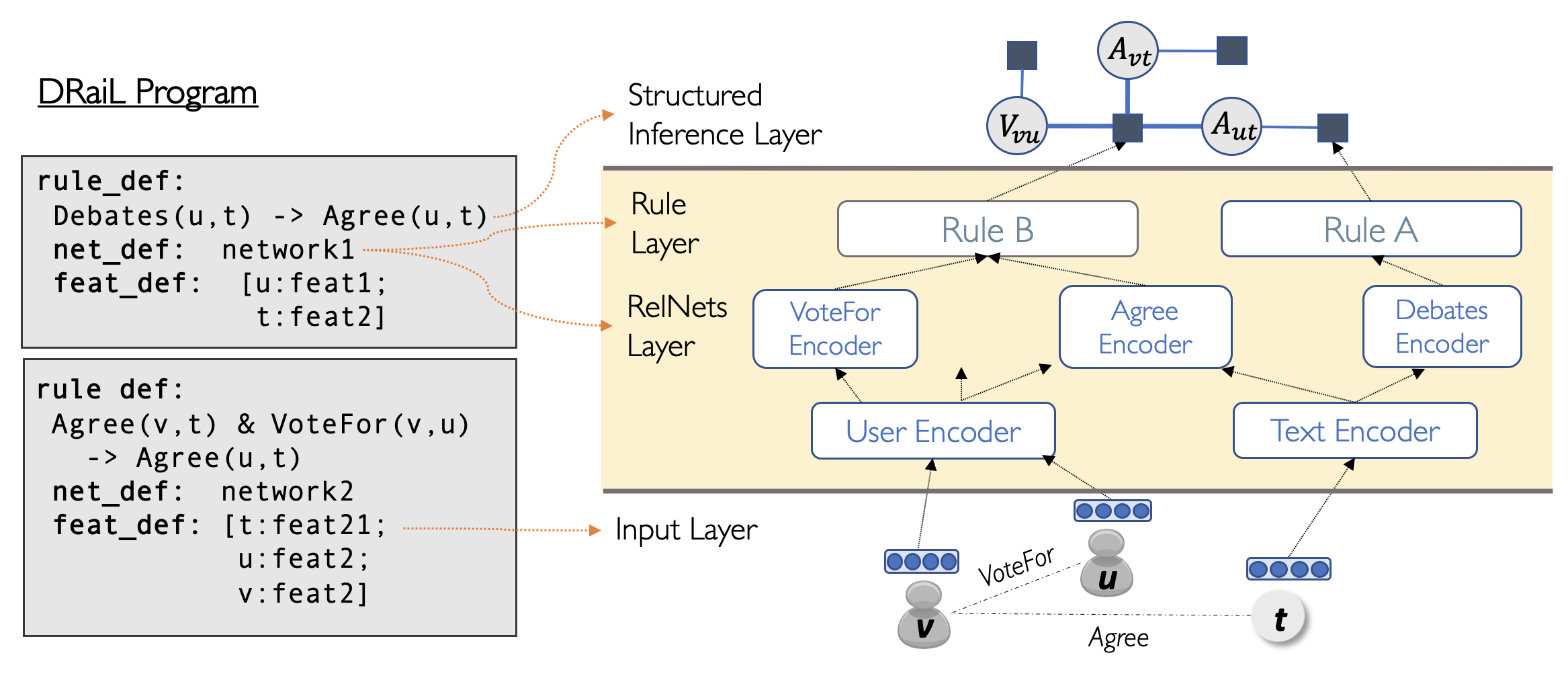}%
    \caption{General overview of \DRAIL}%
    \label{fig:generl_drail}%
\end{figure}

A \DRAIL task is defined by specifying a finite set of \textit{entities} and \textit{relations}. Entities are either discrete symbols (e.g., POS tags, ideologies, specific issue stances), or attributed elements with complex internal information (e.g., documents, users). Decisions are defined using rule \textit{templates}, formatted as horn clauses: $t_{LH} \Rightarrow t_{RH}$, where $t_{LH}$ (\textit{body}) is a conjunction of observed and predicted relations, and $t_{RH}$ (\textit{head}) is the output relation to be learned.
Consider the debate prediction task in Fig.~\ref{fig:exampleDebate}, it consists of several sub-tasks, involving textual inference ({\footnotesize\PRED{Agree(t_1,t_2)}}), social relations ({\footnotesize\PRED{VoteFor(u,v)}}) and their combination  ({\footnotesize\PRED{Agree(u,t)}}). We illustrate how to specify the task as a \DRAIL program in Fig.~\ref{fig:generl_drail} (left), by defining a subset of rule templates to predict these relations.

Each rule template is associated with a neural architecture and a feature function, mapping the initial observations to an input vector for each neural net. We use a shared \textit{relational} embedding space, denoted \relnets, to represent entities and relations over them. As described in Fig.~\ref{fig:generl_drail} (``RelNets Layer''), each entity and relation type is associated with an encoder, trained jointly across all prediction rules. This is a form of relational multi-task learning, as the same entities and relations are reused in multiple rules and their representation is updated accordingly. Each rule defines a neural net, learned over the relations defined on the body. They they take a composition of the vectors generated by the relations encoders as an input (Fig.~\ref{fig:generl_drail}, ``Rule Layer''). \DRAIL is architecture-agnostic, and neural modules for entities, relations and rules can be specified using Pytorch (code snippets can be observed in Appendix \ref{appendix:code}). Our experiments show that we can use different architectures for representing text, users, as well as for embedding discrete entities.

The relations in the horn clauses can correspond to hidden or observed information, and a specific input is defined by the instantiations -or \textit{groundings}- of these elements. 
The collection of all rule groundings results in a factor graph representing our global decision, taking into account the consistency and dependencies between the rules. This way, the final assignments can be obtained by running an inference procedure. 
 For example, the dependency between the users' views on the debate topic (\PRED{Agree(u,t)}) and agreement between them on the topic (\PRED{VoteFor(u,v)}), is modeled as a factor graph in Fig.~\ref{fig:generl_drail} (``Structured Inference Layer'')). 

 We formalize the \DRAIL language in Sec.~\ref{sec:modeling}. Then, in sections \ref{sec:neural}, \ref{sec:relnets} and \ref{sec:learning}, we describe the neural components and learning procedures. 
 
\subsection{Modeling Language}\label{sec:modeling}

We begin our description of \DRAIL by defining the templating language, consisting of entities, relations and rules, and explaining how these elements are instantiated given relevant data.

\textbf{Entities} are named \textit{symbolic} or \textit{attributed} elements. An example of a symbolic entity is a political ideology (e.g. Liberal or Conservative). An example of an attributed entity is a user with age, gender and other profile information, or a document associated with textual content. In \DRAIL entities  can appear either as \textit{constants}, written as strings in double or single quote (e.g. $\mathtt{"user1"}$) or as \textit{variables}, which are identifiers, substituted with constants when grounded. Variables are written using unquoted upper case strings (e.g. $\mathtt{X, X1}$).  Both constants and variables are typed.

\textbf{Relations} are defined between entities and their properties, or other entities. Relations are defined using a unique identifier, a named \textit{predicate}, and a list of typed arguments.  
\textbf{Atoms} consist of a predicate name and a sequence of entities, consistent with the type and arity of the relation's argument list. If the atom's arguments are all constants, it is referred to as a \textbf{ground atom}. For example, {\small\PRED{Agree("user1","user2")}} is a ground atom representing whether {\small$\mathtt{"user1"}$ and $\mathtt{"user2"}$} are in agreement.  When atoms are not grounded (e.g. {\small\PRED{Agree(X,Y)}}) they serve as placeholders for all the possible groundings that can be obtained by replacing the variables with constants. Relations can either be \textit{closed} (i.e., all of their atoms are observed) or \textit{open}, when some of the atoms can be unobserved. In \DRAIL, we use a question mark $\texttt{?}$ to denote unobserved relations. These relations are the units that we reason over. 

To help make these concepts concrete, consider the following example analyzing stances in a debate, as introduced in Fig \ref{fig:exampleDebate}. First, we define the entities.
{\small$\mathtt{User =\{"u1", "u2"\}}, \mathtt{Claim = \{"t1"\}}$
$\mathtt{Post = \{"p1", "p2"\}}$}. Users are entities associated with demographic attributes and preferences. Claims are assertions over which users debate. Posts are textual arguments that users write to explain their position w.r.t the claim. We create these associations by defining a set of relations, capturing authorship {\small$\mathtt{Author(User, Post)}$}, votes between users {\small$\mathtt{VoteFor(User, User)?}$}, and the position users, and their posts, take w.r.t to the debate claim. {\small$\mathtt{Agree(Claim, User)?},~\mathtt{Agree(Claim, Post)?}$}. The authorship relation is the only closed one, e.g., the atom: 
{\small $\mathcal{O}=\{\mathtt{Author("u1","p1")}\}$}.

\textbf{Rules} are functions  that  map  literals (atoms or their negation)  to  other literals. Rules in \DRAIL are defined using templates formatted as horn clauses: $t_{LH} \Rightarrow t_{RH}$, where $t_{LH}$ (\textit{body}) is a conjunction of literals, and $t_{RH}$ (\textit{head}) is the output literal to be predicted, and can only be an instance of open relations. Horn clauses allow us to describe structural dependencies as a collection of ``if-then'' rules, which can be easily interpreted. For example, {\small$\mathtt{Agree(X, C)\wedge VoteFor(Y, X)\Rightarrow Agree(Y,C)}$} expresses the dependency between votes and users holding similar stances on a specific claim. 
We note that rules can be rewritten in \textit{disjunctive form} by converting the logical implication into a disjunction between the negation of the body and the head. For example, the rule above can be rewritten as {\small$\mathtt{\neg Agree(X, C)\vee \neg VoteFor(Y, X)\vee Agree(Y,C)}$}.

\textbf{The \DRAIL program} consists of a set of rules, which can be weighted (i.e., soft constraints), or unweighted (i.e., hard constraints).
Each \textit{weighted rule} template defines a learning problem, used to score assignments to the head of the rule. Since the body may contain open atoms, each rule represents a factor function expressing dependencies between open atoms in the body and head. Unweighted rules, or \textit{constraints}, shape the space of feasible assignments to open atoms, and represent background knowledge about the domain. 

Given the set of grounded atoms $\mathcal{O}$,  rules can be grounded by substituting their variables with constants, such that the grounded atoms correspond to elements in $\mathcal{O}$. This process results in a set of grounded rules, each corresponding to a potential function or to a constraint. Together they define a factor graph. Then, \DRAIL finds the optimally scored assignments for open atoms by performing MAP inference. To formalize this process, we first make the observation that rule groundings can be written as linear inequalities, directly corresponding to their disjunctive form, as follows --
 \begin{equation}\small
  {\small\sum_{i\in I_r^+} y_i + \sum_{i\in I_r^-} (1-y_i) \geq 1}
\label{eq:disjunctive_form}
\end{equation}
\noindent Where $I_r^+$ ($I_r^-$) correspond to the set of open atoms appearing in the rule that are not negated (respectively, negated).
Now, MAP inference can be defined as a linear program. Each rule grounding $r$, generated from template $t(r)$, with input features $\bm{x_r}$ and open atoms $\bm{y_r}$ defines the potential --

\begin{equation}\small
 {\small\psi_{r}(\bm{x_r,y_r})= \min \Big\{\sum_{i\in I_r^+} y_i + \sum_{i\in I_r^-} (1-y_i),  1 \Big\}}
 \label{eq:potentials}
\end{equation}
\noindent added to the linear program with a weight $w_{r}$. Unweighted rule groundings are defined as --
 {\small
\begin{equation}
 c(\mathbf{x_c,y_c})= 1- \sum_{i\in I_c^+} y_i - \sum_{i\in I_c^-} (1-y_i)
\label{eq:constraints}
\end{equation}}
\noindent with $c(\bm{x_c,y_c}) \leq 0$ added as a constraints to the linear program. This way, the MAP problem can be defined over the set of all potentials $\Psi$ and the set of all constraints $C$ as --
\begin{equation}\small
\begin{split}
\argmax_{\bm{y}\in \{0,1\}^n} P(\bm{y}|\bm{x}) \equiv &\argmax_{\bm{y}\in \{0,1\}^n} \sum_{\psi_{r,t} \in \Psi} w_{r}~ \psi_{r}(\bm{x_r,y_r})\notag\\
&\textit{s.t.}~c(\bm{x_c,y_c})\leq0\notag ; \;\;\forall c \in C
\end{split}\label{eq:inference}
\end{equation}
In addition to logical constraints, we also support arithmetic constraints than can be written in the form of linear combinations of atoms with an inequality or an equality. For example, we can enforce the mutual exclusivity of liberal and conservative ideologies for any user $\mathtt{X}$ by writing:
{\small
$$\mathtt{Ideology(X, "con") + Ideology(X, "lib") = 1}$$}
We borrow some additional syntax from PSL to make arithmetic rules easier to use. \citet{bac:jmlr17} define a \textit{summation atom} as an atom that takes terms and/or sum variables as arguments. A summation atom represents the summations of ground atoms that can be obtained by substituting individual variables and summing over all possible constants for sum variables. For example, we could rewrite the above ideology constraint as {\small
$\mathtt{Ideology(X, +I) = 1 }$}. Where {\small
$\mathtt{Ideology(X, +I)}$} represents the summation of all atoms with predicate {\small $\mathtt{Ideology}$} that share variable {\small $\mathtt{X}$}

\DRAIL uses two solvers, Gurobi \cite{gurobi} and AD3 \cite{Martins:2015:AAD:2789272.2789288} for exact and approximate inference respectively.

To ground \DRAIL programs in data, we create an in-memory database consisting of all relations expressed in the program. Observations associated with each relations are provided in column separated text files. \DRAIL's compiler instantiates the program by automatically querying the database and grounding the formatted rules and constraints.

\subsection{Neural Components}\label{sec:neural}

Let $r$ be a rule \textit{grounding} generated from \textit{template} $t$, where $t$ is tied to a neural scoring function $\Phi_{t}$ and a set of parameters $\theta_t$ (Rule Layer in Fig \ref{fig:generl_drail}). In the previous section, we defined the MAP problem for all potentials $\psi_r(\bm{x,y}) \in \Psi$ in a \DRAIL program, where each potential has a weight $w_r$. Consider the following scoring function:
\begin{equation}\small
 w_{r} = \Phi_{t}(\bm{x_r,y_r}; \theta^{t}) = \Phi_t (x_{\text{rel}_0}, ..., x_{\text{rel}_{n-1}}; \theta^t)
\label{eq:rule_composition}
\end{equation}
Notice that all potentials generated by the same template share parameters. We define each scoring function $\Phi_t$ over the set of atoms on the left hand side of the rule template. Let $t = \text{rel}_0 \wedge \text{rel}_1 \wedge ... \wedge \text{rel}_{n-1} \Rightarrow \text{rel}_n$ be a rule template. Each atom $\text{rel}_i$ is composed of a relation type, its arguments and feature vectors for them, as shown in Fig.~\ref{fig:generl_drail}, "Input Layer". 

Given that a \DRAIL program is composed of many competing rules over the same problem, we want to be able to share information between the different decision functions, beyond the probabilistic inference defined in Section~\ref{sec:modeling}. For this purpose, we introduce \textsc{RelNets}.

\subsection{\relnets}\label{sec:relnets}

A \DRAIL program often uses the same entities and relations in multiple different rules. The symbolic aspect of \DRAIL allows us to constrain the values of open relations, and force consistency across all their occurrences. The neural aspect, as defined in Eq. \ref{eq:rule_composition}, associates a neural architecture with each rule template, which can be viewed as a way to embed the output relation. 

We want to exploit the fact that there are repeating occurrences of entities and relations across different rules. Given that each rule defines a learning problem, sharing parameters allows us to shape the representations using complementary learning objectives. This form of relational multi-task learning is illustrated it in Fig.~\ref{fig:generl_drail}, "RelNets Layer".

We formalize this idea by introducing relation-specific and entity-specific encoders and their parameters  $(\phi_{\text{rel}};\theta^{\text{rel}})$ and $(\phi_{\text{ent}}; \theta^{\text{ent}})$, which are reused in all rules. As an example, let's write the formulation for the rules outlined in Fig. \ref{fig:generl_drail}, where each relation and entity encoder is defined over the set of relevant features. 
\begin{equation*}
\begin{split}
 w_{r_0} & = \Phi_{t_0} (\phi_{\text{debates}}(\phi_{\text{user}},\phi_{\text{text}})) \\
 w_{r_1} & = \Phi_{t_1}(\phi_{\text{agree}}(\phi_{\text{user}},\phi_{\text{text}}),\phi_{\text{votefor}}(\phi_{\text{user}},\phi_{\text{user}}))
\end{split}
\label{eq:relnets}
\end{equation*}
Note that entity and relation encoders can be arbitrarily complex, depending on the application. For example, when dealing with text, we could use BiLSTMs or a BERT encoder. 


Our goal when using \relnets is to learn entity representations that capture properties unique to their types (e.g. users, issues), as well as relational patterns that contextualize entities, allowing them to generalize better. We make the distinction between \textit{raw} (or \textit{attributed}) entities and \textit{symbolic} entities. Raw entities are associated with rich, yet unstructured information and attributes, such as text or user profiles. On the other hand, symbolic entities are well defined concepts, and are not associated with additional information, such as political ideologies (e.g. \textit{liberal}) and issues (e.g. \textit{gun-control}). With this consideration, we identify two types of representation learning objectives:

\textbf{Embed Symbol / Explain Data:} Aligns the embedding of symbolic entities and raw entities, grounding the symbol in the raw data, and using the symbol embedding to explain properties of previously unseen raw-entity instances. For example, aligning ideologies and text to (1) obtain an ideology embedding that is closest to the statements made by people with that ideology, or (2) interpret text by providing a symbolic label for it. 



\textbf{Translate / Correlate:} Aligns the representation of pairs of symbolic or raw entities. For example, aligning user representations with text, to move between social and textual information, as shown in Fig. \ref{fig:exampleDebate}, ``Social-Linguistic Relations". Or capturing the correlation between symbolic judgements like agreement and matching ideologies. 

%% file: 4learning.tex
\section{Learning}\label{sec:learning}

The scoring function used for comparing output assignments can be learned \textit{locally} for each rule separately, or \textit{globally}, by considering the dependencies between rules.

\paragraph{Global Learning} The global approach uses inference to ensure that the networks' parameters for all weighted rule templates are consistent across all decisions. Let $\Psi$ be a factor graph with potentials $\{ \psi_r \} \in  \Psi$ over the all possible structures $Y$. Let $\bm{\theta} = \{ \theta^t\}$ be a set of parameter vectors, and $\Phi_{t}(\bm{x_r,y_r}; \theta^t)$ be the scoring function defined for potential $\psi_r(\bm{x_r,y_r})$. Here $\hat{\textbf{y}} \in Y$ corresponds to the current prediction resulting from the MAP inference procedure and $\textbf{y} \in Y$ corresponds to the gold structure. We support two ways to learn $\bm{\theta}$: \\

\noindent (1) The structured hinge loss
{\small
\begin{equation}
\begin{split}
\max(0, \max_{\hat{\textbf{y}} \in Y}(\Delta(\hat{\textbf{y}},\textbf{y}) + \sum_{\psi_r \in \Psi} \Phi_t(\bm{x_r,\hat{y}_r}; \theta^{t})) \\
- \sum_{\psi_r \in \Psi} \Phi_t(\bm{x_r,y_r}; \theta^{t})
\end{split}
\label{eq:structured_hinge_loss}
\end{equation}}
\noindent (2) The general CRF loss
{\footnotesize
\begin{equation} \label{eq:crf_loss}
\begin{split}
 - \text{log} \; p(\textbf{y}|\textbf{x}) &= - \text{log}  \left( \frac{1}{Z(\textbf{x})}   \prod_{\psi_r \in \Psi} \text{exp} \left \{   \Phi_t(\bm{x_r,y_r}; \theta^{t})  \right \} \right) \\
       &= - \sum_{\psi_r \in \Psi} \Phi_t(\bm{x_r,y_r}; \theta^{t}) + \text{log}\; Z(\textbf{x})
\end{split}
\end{equation}}
\noindent Where $Z(\textbf{x})$ is a global normalization term computed over the set of all valid structures $Y$. 
{\footnotesize
\begin{equation*} \label{eq:global_norm}
\begin{split}
Z(\textbf{x}) &= \sum_{\textbf{y'} \in Y} \prod_{\psi_r \in \Psi} \text{exp} \left \{  \Phi_t(\bm{x_r,y'_r}; \theta^{t}) \right \}
\end{split}
\end{equation*}}

When inference is intractable, approximate inference (e.g. AD$^3$) can be used to obtain $\hat{\bm{y}}$. To approximate the global normalization term $Z(\bm{x})$ in the general CRF case, we follow \citet{zhou-etal-2015-neural, Andor2016} and keep a pool $\beta_k$ of $k$ of high-quality feasible solutions during inference. This way, we can sum over the solutions in the pool to approximate the partition function $\sum_{\textbf{y'} \in \beta_k} \prod_{\psi_r \in \Psi} \text{exp} \left \{  \Phi_t(\bm{x_r,y'_r}; \theta^{t}) \right \}$. 

In this paper, we use the structured hinge loss  for most experiments, and include a discussion on the approximated CRF loss in Section \ref{sec:loss_discussion}.

\paragraph{Joint Inference}
The parameters for each weighted rule template are optimized independently. Following \citet{Andor2016}, we show that joint inference serves as a way to greedily approximate the CRF loss, where we replace the normalization term in Eq. \ref{eq:crf_loss} with a greedy approximation over local normalization as:

{\footnotesize
\begin{equation} \label{eq:joint_loss}
\begin{split}
& -\text{log}  \left( \frac{1}{\prod_{\psi_r \in \Psi} Z_L(\bm{x_r})}   \prod_{\psi_r \in \Psi} \text{exp} \left \{    \Phi_{t}(\bm{x_r,y_r}; \theta^{t})   \right \} \right) \\  
 &= - \sum_{\psi_r \in \Psi} \Phi_{t}(\bm{x_r,y_r}; \theta^{t}) + \sum_{\psi_r \in \Psi} \text{log} \; Z_L(\bm{x_r})
\end{split}
\end{equation}}

\noindent where $Z_L(\bm{x_r})$ is computed over all the valid assignments $\bm{y'_r}$ for each factor $\psi_{r}$. We refer to models that use this approach as \Inference.
{\footnotesize
\begin{equation*} \label{eq:local_norm}
\begin{split}
Z_L(\textbf{x}_{r}) &= \sum_{\bm{y'_r}} \text{exp} \left \{  \Phi_{t}(\bm{x_r,y'_r}; \theta^{t}) \right \}
\end{split}
\end{equation*}}

\vspace{-.5 cm}

%% file: 7experiments.tex
\section{Experimental Evaluation}\label{sec:experiments}


We compare \DRAIL to representative models from each category covered in Sec \ref{sec:related}. 
Our goal is to examine how different types of approaches capture dependencies and what are their limitations when dealing with language interactions. These baselines are described in Sec. \ref{sec:baselines}. We also evaluate different strategies using \DRAIL in Sec. \ref{sec:strategies}.

We focus on three tasks: open debate stance prediction (Sec. \ref{sec:open_debate}), issue-specific stance prediction (Sec. \ref{sec:4forums}) and argumentation mining (Sec. \ref{sec:arg_mining}). Details regarding the hyper-parameters used for all tasks can be found in Appendix \ref{appendix:hyperparams}.

\subsection{Baselines}\label{sec:baselines}

\textbf{End-to-end Neural Nets:} We test all approaches against neural nets trained locally on each task, without explicitly modeling dependencies. In this space, we consider two variants: \Independent, where each component of the problem is represented using an independent neural network, and \Endtoend, where the features for the different components are concatenated at the input and fed to a single neural network. 

\textbf{Relational Embedding Methods:} Introduced in Sec. \ref{sec:related_node_embedding}, these methods embed nodes and edge types for relational data. Typically designed to represent symbolic entities and relations, however since our entities can be defined by raw textual content and other features, we define the relational objectives over our encoders. This adaptation has proven successful for domains dealing with rich textual information \cite{lee-goldwasser-2019-multi}. We test three relational knowledge objectives: TransE \cite{transE}, ComplEx \cite{pmlr-v48-trouillon16} and RotatE \cite{sun2018rotate}.
\textbf{Limitations:} (1) These approaches cannot constrain the space using domain knowledge, and (2) they cannot deal with relations involving more than two entities, limiting their applicability to higher order factors. 

\textbf{Probabilistic Logics:} We compare to PSL \cite{bac:jmlr17}, a purely symbolic probabilistic logic, and TensorLog \cite{DBLP:journals/jair/CohenYM20}, a neuro-symbolic one. In both cases, we instantiate the program using the weights learned with our base encoders. \textbf{Limitations:} These approaches do not provide a way to update the parameters of the base classifiers. 

\subsection{Modeling Strategies}\label{sec:strategies}



\textbf{Local vs. Global Learning:} The trade-off between local and global learning has been explored for graphical models (MEMM vs. CRF), and for deep structured prediction~ \cite{chen2014fast,Andor2016,han-etal-2019-joint}. While local learning is faster, the learned scoring functions might not be consistent with the correct global prediction. Following \citealt{han-etal-2019-joint}, we initialize the parameters using local models. 



\textbf{\relnets:} We will show the advantage of having relational representations that are shared across different decisions, in contrast to having independent parameters for each rule. Note that in all cases, we will use the global learning objective to train \relnets.  

\textbf{Modularity:} Decomposing decisions into relevant modules has been shown to simplify the learning process and lead to better generalization \cite{zhang-goldwasser-2019-sentiment}. We will contrast the performance of \textit{modular} and \textit{end-to-end} models to represent text and user information when predicting stances.

\textbf{Representation Learning and Interpretability:} We will do a qualitative analysis to show how we are able to \textit{embed symbols} and \textit{explain data} by moving between symbolic and sub-symbolic representations, as outlined in Section \ref{sec:relnets}. 


\begin{table*}[!htb]
     \begin{minipage}{0.5\linewidth}
      \centering
        \resizebox{0.80\linewidth}{!}{%
            \begin{tabular}{llccc|ccc}
            \toprule
        & \textbf{Model} & \multicolumn{3}{c}{\textbf{Random}} & \multicolumn{3}{c}{\textbf{Hard}} \\
        \midrule
        ~ & ~ & \textbf{P} & \textbf{U} & \textbf{V} & \textbf{P} & \textbf{U} & \textbf{V}\\
        \midrule
        \multirow{2}{*}{\bf Local} 
            & \Independent & 63.9 & 61.3 & 54.4 & 62.2 & 53.0 & 51.3 \\
            & \Endtoend & 66.3 & 71.2 & 54.4 & 63.4 & 68.1 & 51.3   \\
        \midrule
        {\bf Reln.}
            & TransE  & 58.5 & 54.1 & 52.6 & 57.2 & 53.1 & 51.2  \\
        {\bf Emb.} 
            & ComplEx  & 61.0 & 63.3 & \textbf{58.1} & 57.3 & 55.0 & 55.4  \\
            & RotatE    & 59.6 & 58.3 & 54.2 & 57.9 & 54.6 & 51.0  \\
        \midrule
        {\bf Prob.}
            & PSL & 78.7 & 77.5 & 55.4 & 72.6 & 71.8 & 52.6 \\
        {\bf Logic.}    
            & TensorLog  & 72.7 & 71.9 & 56.2 & 70.0 & 67.4 & \textbf{55.8}  \\
        \midrule
        \multirow{4}{*}{\bf DRaiL} 
            & \Endtoend+Inf & 80.2 & 79.2 &  54.4 & 76.9 & 75.5 & 51.3 \\
            & \Inference  & 80.7 & 79.5 & 55.6 & 75.2 & 74.0 & 52.5  \\
            & \Global & 81.0 & 79.5 & 55.8 &  75.3 & 74.0 & 53.0  \\
            & \glrelnets & \textbf{81.9} & \textbf{80.4} & 57.0 & \textbf{78.0} & \textbf{77.2} & 53.7 \\
        \bottomrule   
    \end{tabular}}
    \end{minipage}%
    \begin{minipage}{0.5\linewidth}
      \centering
        \resizebox{0.80\linewidth}{!}{%
          \begin{tabular}{llccc|ccc}
            \toprule
            \ & \textbf{Model} & \multicolumn{3}{c}{\textbf{Random}} & \multicolumn{3}{c}{\textbf{Hard}} \\
            \midrule
            ~ & ~ & \textbf{P} & \textbf{U} & \textbf{V} & \textbf{P} & \textbf{U} & \textbf{V} \\
            \midrule
            \multirow{2}{*}{\bf Local} 
             & \Independent & 63.9 & 61.3 & 54.4 & 62.2 & 53.0 & 51.3 \\
             & \Endtoend & 66.3 & 71.2 & 54.4 & 63.4 & 68.1 & 51.3   \\
            \midrule
            \multirow{3}{*}{\bf AC}
            & \Inference & 73.6 & 71.8 & - & 69.0 & 67.2 & -  \\
            & \Global  & 73.6 & 72.0 & - & 69.0 & 67.2 & -  \\
            & \glrelnets & 73.8 & 72.0 & - & 71.7 & 69.5 & -  \\
           \midrule
        \bf{AC}
            & \Inference & 80.7 & 79.5 & - & 75.6 & 74.4 & - \\
        \bf{DC}
            & \Global & 81.4 & 79.9 & - & 75.8 & 74.6 & -\\
            & \glrelnets & 81.8 & 80.1 & - & 77.8 & 76.4 & -  \\
            \midrule
        \bf{AC}
            & \Inference & 80.7 & 79.5 & 55.6 & 75.2 & 74.0 & 52.5  \\
        \bf{DC}
            & \Global & 81.0  & 79.5 & 55.8 & 75.3 & 74.0 & 53.0  \\
        \bf{SC}
            & \glrelnets  & \textbf{81.9} & \textbf{80.4} & \textbf{57.0} & \textbf{78.0} & \textbf{77.2} & \textbf{53.7}  \\
            \bottomrule
        \end{tabular}}
    \end{minipage}
    \centering
    \caption{\small{General Results for Open Domain Stance Prediction (Left), Variations of the Model (Right). \\ P:Post, U:User, V:Voter}}\label{tab:open_general}
\end{table*}

\subsection{Open Domain Stance Prediction}\label{sec:open_debate}

Traditionally, stance prediction tasks have focused on predicting stances on a specific topic, such as abortion. Predicting stances for a different topic, such as gun control would require learning a new model from scratch. In this task, we would like to leverage the fact that stances in different domains are correlated. Instead of using a pre-defined set of debate topics (i.e., \textit{symbolic} entities) we define the prediction task over claims, expressed in text, specific to each debate. Concretely, each debate will have a different claim (i.e., different value for  {\small$\mathtt{C}$} in the relation  {\small$\mathtt{Claim(T,C)}$}, where $\mathtt{T}$ corresponds to a debate thread). We refer to these settings as \textit{Open-Domain} and write down the task in Fig. \ref{fig:open_rules}. In addition to the textual stance prediction problem (r0), where $\mathtt{P}$ corresponds to a post, we represent users ($\mathtt{U}$) and define a user-level stance prediction problem (r1). We assume that additional users read the posts and vote for content that supports their views, resulting in another prediction problem (r2,r3). Then, we define representation learning tasks, which align symbolic (ideology, defined as $\mathtt{I}$) and raw (users and text) entities (r4-r7). Finally, we write down all dependencies and constrain the final prediction (c0-c7).

\begin{figure}[h!]
    \centering
    \scriptsize
    \begin{tabular}{l}
    \hline
        \textit{// Predictions} \\
        \textbf{r0:} \PRED{InThread(T,P)} $\wedge$ \PRED{Claim(T,C)} $\Rightarrow$ \PRED{Agree(P,C)}?\\
        
        \textbf{r1:} \PRED{Debates(T,U)} $\wedge$ \PRED{Claim(T,C)} $\Rightarrow$ \PRED{Agree(U,C)}?\\ 
        
        \textbf{r2:} \PRED{Debates(T,U)} $\wedge$ \PRED{Votes(T,V)} $\Rightarrow$ \PRED{VoteFor(V,U)}?\\ 
        
        \textbf{r3:} \PRED{Votes(T,V_1)}  $\wedge$ \PRED{Votes(T,V_2)} $\Rightarrow$ \PRED{VoteSame(V_1,V_2)?}\\ 
        
        \textit{// Auxiliary objectives} \\
        \textbf{r4:} \PRED{InThread(T,P)} $\wedge$ \PRED{Ideology(I)} $\Rightarrow$ \PRED{HasIdeology(P,I)?} \\
        \textbf{r5:} \PRED{Claim(T,C)} $\wedge$ \PRED{Ideology(I)} $\Rightarrow$ \PRED{HasIdeology(C,I)?} \\
        \textbf{r6:} \PRED{Debates(T,U)} $\wedge$ \PRED{Ideology(I)} $\Rightarrow$ \PRED{HasIdeology(U,I)?} \\
        \textbf{r7:} \PRED{HasIdeology(A,I)?} $\wedge$ \PRED{HasIdeology(B,I)?} $\Rightarrow$ \PRED{Agree(A,B)}?\\
        \\
        
         \textit{// Author constraints} \\
        \textbf{c0:} \PRED{Agree(P,C)}? $\wedge$ \PRED{Author(P,U)} $\Rightarrow$ \PRED{Agree(U,C)}?\\ 
        \textbf{c1:} \NPRED{Agree(P,C)}? $\wedge$ \PRED{Author(P,U)} $\Rightarrow$ \NPRED{Agree(U,C)}?\\ 
         \textit{// Debate constraints} \\
        \textbf{c2:} \PRED{Agree(P_1,C)}? $\wedge$ \PRED{Respond(P_1,P_2)} $\Rightarrow$ \NPRED{Agree(P_2,C)}?\\ 
        \textbf{c3:} \NPRED{Agree(P_1,C)}? $\wedge$ \PRED{Respond(P_1,P_2)} $\Rightarrow$ \PRED{Agree(P_2,C)}?\\ 
         \textit{// Social constraints} \\
        \textbf{c4:} \PRED{Agree(U,C)}? $\wedge$ \PRED{VoteFor(V,U)}? $\Rightarrow$ \PRED{Agree(V,U)}?\\
        \textbf{c5:} \NPRED{Agree(U,C)}? $\wedge$ \PRED{VoteFor(V,U)}? $\Rightarrow$ \NPRED{Agree(V,U)}?\\
        \textbf{c6:} \PRED{Agree(V_1,C)}? $\wedge$ \PRED{VoteSame(V_1,V_2)}? $\Rightarrow$ \PRED{Agree(V_2,C)}?\\
        \textbf{c7:} \NPRED{Agree(V_1,C)}? $\wedge$ \PRED{VoteSame(V_1,V_2)}? $\Rightarrow$ \NPRED{Agree(V_2,C)}?\\

    \hline
    \end{tabular}\caption{{\small \DRAIL Program for O.D. Stance Prediction.}\\
     \scriptsize{T: Thread, C: Claim, P: Post, U: User, V: Voter, I: Ideology, A,B: Can be any in \{Claim, Post, User\}}}
    \label{fig:open_rules}
\end{figure}

\textbf{Dataset:} We collected a set of 7,555 debates from \url{debate.org}, containing a total of 42,245 posts across 10 broader political issues. For a given issue, the debate topics are nuanced and vary according to the debate question expressed in text (e.g. \textit{Should semi-automatic guns be banned}, \textit{Conceal handgun laws reduce violent crime}). Debates have at least two posts, containing up to 25 sentences each. In addition to debates and posts, we collected the user profiles of all users participating in the debates, as well as all users that cast votes for the debate participants. Profiles consist of attributes (e.g. gender, ideology). User data is considerably sparse. 
We create two evaluation scenarios, \textit{random} and \textit{hard}. In the random split, debates are randomly divided into ten folds of equal size. In the hard split, debates are separated by political issue. This results in a harder prediction problem, as the test data will not share topically related debates with the training data. We perform 10-fold cross validation and report accuracy.

\textbf{Entity and Relation Encoders:} 
We represent posts and titles using a pre-trained BERT-small\footnote{We found negligible difference in performance between BERT and BERT-small for this task, while obtaining a considerable boost in speed} encoder~\cite{turc2019}, a compact version of the language model proposed by \citealt{devlin2018bert}. For users, we use feed-forward computations with ReLU activations over the profile features and a pre-trained node embedding \cite{node2vec-kdd2016} over the friendship graph. All relation and rule encoders are represented as feed-forward networks with one hidden layer, ReLU activations and a softmax on top. Note that all of these modules are updated during learning. 

\begin{table*}[ht]
    \centering
    \resizebox{\linewidth}{!}{%
        \begin{tabular}{l|llllll}
        \toprule
        \textbf{Issue} & \textbf{Debate Statements }&\textbf{ Con}\ &\textbf{ Libt} & \textbf{Mod} & \textbf{Libl} & \textbf{ Pro} \\
        \midrule
        \multirow{5}{*}{\textbf{Guns}} 
        & No gun laws should be passed restricting the right to bear arms & \textbf{.98} & .01 & .00 & .01 & .00 \\
        & Gun control is an ineffective comfort tactic used by the government to fool the American people & .08 & \textbf{.65} & .22 & .02 & .03 \\
        & Gun control is good for society & .14 & .06 & \textbf{.60} & .15 & .06 \\ 
        & In the US handguns ought to be banned & .03 & .01 & .01 & \textbf{.93} & .02 \\
        & The USA should ban most guns and confiscate them & .00 & .00 & .01 & .00 & \textbf{.99}\\
        \bottomrule
        \end{tabular}
        }
    \newline
    \vspace{.5cm}
    \newline
    \resizebox{\linewidth}{!}{%
    \begin{tabular}{c|c|l}
    \toprule
    \textbf{Issue} &\textbf{ Ideology} & \textbf{Statements close in the embedding space} \\
    \midrule
    \multirow{2}{*}{\textbf{LGBT}} & \multirow{1}{*}{\textbf{Libl}} & gay marriage ought be legalized, gay marriage should be legalized, same-sex marriage should be federally legal \\
    \cline{2-3}
    ~ & \multirow{1}{*}{\textbf{Con}} & Leviticus 18:22 and 20:13 prove the anti gay marriage position, gay marriage is not bad, homosexuality is not a sin nor taboo \\
    
    \bottomrule
       
    \end{tabular}}

    \caption{Representation Learning Objectives: Explain Data (Top) and Embed Symbol (Bottom).\\ \footnotesize{Note that ideology labels were learned from user profiles, and do not necessarily represent the official stances of political parties.}}\label{tab:interpretability}
    
\end{table*}

\begin{figure*}[ht]%
    \begin{minipage}{0.4\textwidth}
    \input{figs/politicians_guns}
    \end{minipage}
    \begin{minipage}{0.5\textwidth}
    \resizebox{1.2\linewidth}{!}{%
    \begin{tabular}{llll}
    \toprule
        \textbf{Politician} & \textbf{Issue} & \textbf{Statement} & \textbf{Label} \\
    \midrule
         & Guns & For background checks, and closing loopholes & Left \\
        \textbf{Sanders} & Guns & Intervene with mental illness, to prevent mass shootings & Mod \\
         & Abortion & Advocate for family planning and funding for contraceptives & Left \\ 
        
    \midrule
         & Guns & Guns need to have trigger locks & Left \\
        \textbf{Biden} & Abortion & Accepts catholic church view that life begins at conception & Right \\
         & Abortion & Ensure access to and funding for contraception & Left \\
    \midrule
         & Guns & No limits on guns; they save lives & Right\\
        \textbf{Trump} & Abortion & I am pro-life; fight ObamaCare abortion funding & Right \\
        & Abortion & Planned Parenthood does great work on women's health & Left\\
    \bottomrule
        
    \end{tabular}}
    
    \end{minipage}
    \caption{Statements Made by Politicians Classified Using our Model Trained on debate.org.}
	\label{table:ontheissues}
\end{figure*}

Tab. \ref{tab:open_general} (Left) shows results for all the models described in Section \ref{sec:baselines}. In \Endtoend models, post and user information is collapsed into a single module (rule), whereas in \Independent, \Inference, \Global and \relnets they are modeled separately. All other baselines use the same underlying modular encoders. We can appreciate the advantage of relational embeddings in contrast to \Independent for user and voter stances, particularly in the case of ComplEx and RotatE. We can attribute this to the fact that all objectives are trained jointly and entity encoders are shared. However, approaches that explicitly model inference, like PSL, TensorLog and \DRAIL outperform relational embeddings and end-to-end neural networks. This is because they enforce domain constraints.

We explain the difference between the performance of \DRAIL and the other probabilistic logics by: (1) The fact that we use exact inference instead of approximate inference, (2) PSL learns to weight the rules without giving priority to a particular task, whereas the \Inference model works directly over the local outputs, and \textit{most importantly} (3) our \Global and \relnets models back-propagate to the base classifiers and fine-tune parameters using a structured objective.

%

In Tab. \ref{tab:open_general} (Right) we show different versions of the \DRAIL program, by adding or removing certain constraints. AC models only enforce author consistency, AC-DC models enforce both author consistency and disagreement between respondents, and finally, AC-DC-SC models introduce social information by considering voting behavior. We get better performance when we model more contextualizing information for the \relnets case. This is particularly helpful in the \textit{Hard} case, where contextualizing information, combined with shared representations, help the model generalize to previously unobserved topics. With respect to the modeling strategies listed in Section \ref{sec:strategies}, we can observe: (1) The advantage of using a global learning objective, (2) the advantage of using \relnets to share information and (2) the advantage of breaking down the decision into modules, instead of learning an end-to-end model. 

Then, we perform a qualitative evaluation to illustrate our ability to move between symbolic and raw information. Tab. \ref{tab:interpretability} (Top) takes a set of statements and \textit{explains} them by looking at the symbols associated with them and their score. For learning to map debate statements to ideological symbols, we rely on the partial supervision provided by the users that self-identify with a political ideology and disclose it on their public profiles. Note that we do not incorporate any explicit expertise in political science to learn to represent ideological information. We chose statements with the highest score for each of the ideologies. We can see that, in the context of guns, statements that have to do with some form of gun control have higher scores for the center-to-left spectrum of ideological symbols (moderate, liberal, progressive), whereas statements that mention gun rights and the ineffectiveness of gun control policies have higher scores for conservative and libertarian symbols.

To complement this evaluation, in Tab. \ref{tab:interpretability} (Bottom), we \textit{embed} ideologies and find three example statements that are close in the embedding space. In the context of LGBT issues, we find that statements closest to the liberal symbol are those that support the legalization of same-sex marriage, and frame it as a constitutional issue. On the other hand, the statements closest to the conservative symbol, frame homosexuality and same-sex marriage as a moral or religious issue, and we find statements both supporting and opposing same-sex marriage. This experiment shows that our model is easy to interpret, and provides an explanation for the decision made. 

Finally, we evaluate our learned model over entities that have not been observed during training. To do this, we extract statements made by three prominent politicians from \textit{ontheissues.org}. Then, we try to explain the politicians by looking at their predicted ideology. Results for this evaluation can be seen in Tab. \ref{table:ontheissues}. The left Fig. shows the proportion of statements that were identified for each ideology: left (liberal or progressive), moderate and right (conservative). We find that we are able to recover the relative positions in the political spectrum for the evaluated politicians: Bernie Sanders, Joe Biden and Donald Trump. We find that Sanders is the most left leaning, followed by Biden. In contrast, Donald Trump stands mostly on the right. We also include some examples of the classified statements. We show that we are able to identify cases in which the statement does not necessarily align with the known ideology for each politician. 



\subsection{Issue-Specific Stance Prediction}\label{sec:4forums}

Given a debate thread on a specific issue (e.g. abortion), the task is to predict the stance w.r.t. the issue for each one of the debate posts~\cite{Walker:2012:SCU:2382029.2382124}. Each thread forms a tree structure, where  users participate and respond to each other's posts. We treat the task as a collective classification problem, and model the agreement between posts and their replies, as well as the consistency between posts written by the same author. The \DRAIL program for this task can be observed in Appendix \ref{appendix:programs}.

\textbf{Dataset:} We use the 4Forums dataset from the Internet Argument Corpus \cite{Walker:2012:SCU:2382029.2382124}, consisting of a total of 1,230 debates and 24,658 posts on abortion, evolution, gay marriage and gun control. We use the same splits as \citealt{C18-1316} and perform 5-fold cross validation. 

\textbf{Entity and Relation Encoders:} We represented posts using pre-trained BERT encoders \cite{devlin2018bert} and do not generate features for authors. As in the previous task, we model all relations and rules using feed-forward networks with one hidden layer and ReLU activations. Note that we fine-tune all parameters during training.

In Tab. \ref{tab:poststance} we can observe the general results for this task. We report macro F1 for post stance and agreement between posts for all issues. As in the previous task, we find that ComplEx and RotatE relational embeddings outperform \Independent, and probabilistic logics outperform methods that do not perform constrained inference. PSL outperforms \Inference for evolution and gun control debates, which are the two issues with less training data. Whereas \Inference outperforms PSL for debates on abortion and gay marriage. This could indicate that re-weighting rules may be advantageous for the cases with less supervision. Finally, we see the advantage of using a global learning objective and augmenting it with shared representations. Tab. \ref{tab:4forums_previous_work} compares our model with previously published results. 

\begin{table}[t]
    \centering
    \resizebox{\columnwidth}{!}{%
        \begin{tabular}{llcc|cc|cc|cc}
        \toprule
              & \multicolumn{1}{c}{Model}  & \multicolumn{2}{c}{AB} & \multicolumn{2}{c}{E} & \multicolumn{2}{c}{GM} & \multicolumn{2}{c}{GC} \\ 
             \midrule

            \multicolumn{2}{c}{}  & S & A
            & S & A
            & S & A
            & S & A \\
            \midrule
        \multirow{1}{*}{\bf Local} 
            &    \Independent   & 66.0 & 61.7 & 58.2 & 59.7 & 62.6 & 60.6 & 59.5 & 61.0 \\

        \midrule
        {\bf Reln.}
            & TransE & 62.5 & 62.9 & 53.5 & 65.1 & 58.7 & 69.3 & 55.3 & 65.0  \\
        {\bf Embed.} 
            & ComplEx & 66.6 & 73.4 & 60.7 & 72.2 & 66.6 & 72.8 & 60.0 & 70.7  \\
            & RotatE  & 66.6 & 72.3 & 59.2 & 71.3 & 67.0 & 74.2 & 59.4 & 69.9 \\
        \midrule
        {\bf Prob.}
            & PSL & 81.6 & 74.4 & 69.0 & 64.9 & 83.3 & 74.2 & 71.9 & 71.7 \\
        {\bf Logic.}    
            & TensorLog & 77.3 & 61.3 & 68.2 & 51.3 & 80.4 & 65.2 & 68.3 & 55.6\\

        \midrule
        \multirow{3}{*}{\bf \DRAIL}
            & \Inference       & 82.8 & 74.6 & 64.8 & 63.2 & 84.5 & 73.4 & 70.4 & 66.3 \\
            & \Global & 88.6 & \textbf{84.7} & 72.8 & 72.2 & \textbf{90.3} & 81.8 & 76.8 & 72.2\\\ 
            & \glrelnets & \textbf{89.0} & 83.5 & \textbf{80.5} & \textbf{76.4} & 89.3 & \textbf{82.1} & \textbf{80.3} & \textbf{73.4} \\
          
        \bottomrule
        \end{tabular}
    }
    \caption{General Results for Issue-Specific Stance and Agreement Prediction (Macro F1).
    \footnotesize{AB: Abortion, E: Evolution, GM: Gay Marriage, GC: Gun Control}}
    \label{tab:poststance}
\end{table}%

\begin{table}[t]\small
    \centering
    \resizebox{\columnwidth}{!}{%
    \begin{tabular}{lccccc}
        \toprule
        Model & A & E & GM & GC  & Avg\\
        \midrule
        BERT \cite{devlin2018bert} & 67.0 & 62.4 & 67.4 & 64.6 & 65.4 \\
        PSL \cite{sridhar:acl15}   & 77.0 & 80.3 & 80.5 & 69.1 & 76.7\\
        Struct. Rep. \cite{C18-1316} & 86.5 & 82.2 & 87.6 & 83.1 & 84.9\\
        \midrule
        \DRAIL \glrelnets & \textbf{89.2} & \textbf{82.4} & \textbf{90.1} & \textbf{83.1} & \textbf{86.2}  \\
        \bottomrule

    \end{tabular}
    }
    \caption{Previous work on Issue-Specific Stance Prediction (Stance Acc.)}
    \label{tab:4forums_previous_work}
\end{table}

\subsection{Argument Mining}\label{sec:arg_mining}

The goal of this task is to identify argumentative structures in essays. Each argumentative structure corresponds to a tree in a document. Nodes are predefined spans of text and can be labeled either as \textit{claims, major claims} or \textit{premises}, and edges correspond to support/attack relations between nodes. Domain knowledge is injected by constraining sources to be premises and targets to be either premises or major claims, as well as enforcing tree structures. We model nodes, links and second order relations, \textit{grandparent} ($a\rightarrow b\rightarrow c$), and \textit{co-parent} ($a\rightarrow b\leftarrow  c$) \cite{niculae-etal-2017-argument}. Additionally, we consider link labels, denoted stances. The \DRAIL program for this task can be observed in Appendix \ref{appendix:programs}.

\textbf{Dataset:} We used the UKP dataset \cite{Stab2017PAS}, consisting of 402 documents,  with  a  total of 6,100  propositions and 3,800 links (17\% of pairs). We use the splits used by \citeauthor{niculae-etal-2017-argument} \shortcite{niculae-etal-2017-argument}, and report macro F1 for components and positive F1 for relations.

\textbf{Entity and Relation Encoders:} To represent the component and the essay, we used a BiLSTM over the words, initialized with Glove embeddings \cite{pennington2014glove}, concatenated with a feature vector following  \citeauthor{niculae-etal-2017-argument} \shortcite{niculae-etal-2017-argument}. For representing the relation, we use a feed-forward computation over the components, as well as the relation features used in \citeauthor{niculae-etal-2017-argument} \shortcite{niculae-etal-2017-argument}. 

We can observe the general results for this task in Tab. \ref{tab:deep_argument}. Given that this task relies on constructing the tree from scratch, we find that all methods that do not include declarative constraints (\Independent and relational embeddings) suffer when trying to predict links correctly. For this task, we did not apply TensorLog given that we couldn't find a way to express tree constraints using their syntax. Once again, we see the advantage of using global learning, as well as sharing information between rules using \relnets.

\begin{table}[ht]
    \centering
    \resizebox{\columnwidth}{!}{%
        \begin{tabular}{llccccc}
            \toprule
            \multicolumn{1}{l}{} & \multicolumn{1}{l}{Model} & Node & Link & Avg & Stance & Avg \\
            \midrule
            \multirow{1}{*}{\bf Local} 
             & \Independent  & 70.7 & 52.8 & 61.7 & 63.4  & 62.3 \\
            \midrule
            {\bf Reln.}
            & TransE & 65.7 & 23.7 & 44.7 & 44.6 & 44.7 \\
            {\bf Embed.}
            & ComplEx & 69.1 & 15.7 & 42.4 & 53.5 & 46.1 \\
            & RotatE  & 67.2 & 20.7 & 44.0 & 46.7 & 44.9 \\
            \midrule
            {\bf Prob. Logic}
                & PSL & 76.5 & 56.4 & 66.5 & 64.7 & 65.9  \\
           
            \midrule
            \multirow{3}{*}{\bf \DRAIL} 
                & \Inference    & 78.6 & 59.5 & 69.1 & 62.9 & 67.0 \\
                & \Global & \textbf{83.1} & 61.2 & 72.2 & \textbf{69.2} & 71.2  \\
                & \glrelnets & 82.9 & \textbf{63.7} & \textbf{73.3} & 68.4 & \textbf{71.7} \\
            \bottomrule
        \end{tabular}
    }
    \caption{General Results for Argument Mining}
    \label{tab:deep_argument}
\end{table}

Tab. \ref{tab:argument_mining_previous_work} shows the performance of our model against previously published results. While we are able to outperform models that use the same underlying encoders and features, recent work by \citealt{kuribayashi-etal-2019-empirical} further improved performance by exploiting contextualized word embeddings that look at the whole document, and making a distinction between argumentative markers and argumentative components. We did not find a significant improvement by incorporating their ELMo-LSTM encoders into our framework\footnote{We did not experiment with their normalization approach, extended BoW features, nor AC/AM distinction.}, nor by replacing our BiLSTM encoders with BERT. We leave the exploration of an effective way to leverage contextualized embeddings for this task for future work. 

\begin{table}[ht]
    \centering\small
    \resizebox{\columnwidth}{!}{%
    \begin{tabular}{lccc}
    \toprule
    \multicolumn{1}{l}{Model} & Node & Link & Avg \\
    \midrule
    Human upper bound   & 86.8 & 75.5 & 81.2 \\
    \midrule
    ILP Joint \cite{Stab2017PAS} & 82.6 & 58.5 & 70.6 \\ 
    Struct RNN strict  \cite{niculae-etal-2017-argument}  & 79.3 & 50.1 & 64.7 \\
    Struct SVM full \cite{niculae-etal-2017-argument} & 77.6 & 60.1 & 68.9 \\ 
    Joint PointerNet \cite{potash-etal-2017-heres} & 84.9 & 60.8 & 72.9 \\ 
    \citealt{kuribayashi-etal-2019-empirical} & \textbf{85.7} & \textbf{67.8} & \textbf{76.8} \\
       \midrule
    \DRAIL \glrelnets & 82.9 & 63.7 & 73.3  \\
    \bottomrule
    \end{tabular}
    }
    \caption{Previous Work on Argument Mining}
    \label{tab:argument_mining_previous_work}
\end{table}

\subsection{Run-time Analysis}

In this section, we perform a run-time analysis of all probabilistic logic systems tested. 
All experiments were run on a 12 core 3.2Ghz Intel i7 CPU machine with 63GB RAM and an NVIDIA GeForce GTX 1080 Ti 11GB GDDR5X GPU.

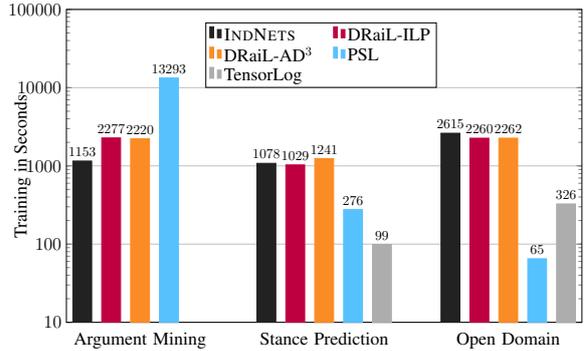
\begin{figure}[h]
    \resizebox{\columnwidth}{!}{%
        \input{figs/training_runtime.tex}
    }
    \caption{Avg. Overall Training Time (per Fold)}
    \label{fig:train-runtime}
\end{figure}

Fig. \ref{fig:train-runtime} shows the overall training time (per fold) in seconds for each of the evaluated tasks. Note that the figure is presented in logarithmic scale. We find that \DRAIL is generally more computationally expensive than both TensorLog and PSL. This is expected given that \DRAIL back-propagates to the base classifiers at each epoch, while the other frameworks just take the local predictions as priors. However, when using a large number of arithmetic constraints (e.g. Argument Mining), we find that PSL takes a really long time to train. 
We found no significant difference when using ILP or AD$^3$. We presume that this is due to the fact that our graphs are small and that Gurobi is a highly optimized commercial software. 

Finally, we find that when using encoders with a large number of parameters (e.g. BERT) in tasks with small graphs, the difference in training time between training local and global models is minimal. In these cases, back-propagation is considerably more expensive than inference, and global models converge in fewer epochs. For Argument Mining, local models are at least twice as fast. BiLSTMs are considerably faster than BERT, and inference is more expensive for this task. 


\subsection{Analysis of Loss Functions}\label{sec:loss_discussion}

In this section we perform an evaluation of the CRF loss for issue-specific stance prediction. Note that one drawback of the CRF loss (Eq. \ref{eq:crf_loss}) is that we need to accumulate the gradient for the approximated partition function. When using entity encoders with a lot of parameters (e.g. BERT), the amount of memory needed for a single instance increases. We were unable to fit the full models in our GPU. For the purpose of these tests, we froze the BERT parameters after local training and updated only the relation and rule parameters. To obtain the solution pool, we use gurobi's pool search mode to find $\beta$ high-quality solutions. This also increases the cost of search at inference time. 

Development set results for the debates on abortion can be observed in Tab. \ref{tab:loss_func}. While increasing the size of the solution pool leads to better performance, it comes at a higher computational cost. 

\begin{table}[h]
    \centering
    \resizebox{\columnwidth}{!}{%
    \begin{tabular}{lcccc}
    \toprule
        Model & Stance & Agree & Avg & Secs per epoch \\
        \midrule
        Hinge loss & 82.74 &	78.54 &	80.64 &	132 \\
        \midrule
        CRF($\beta=5$) & 83.09 & 81.03 & 82.06 &	345 \\
        CRF($\beta=20$) & 84.10 &	82.16 &	83.13 & 482 \\
        CRF($\beta=50$) & 84.19	& 81.80 &	83.00 & 720 \\
        \bottomrule
    \end{tabular}
    }
    \caption{Stance Prediction (Abortion) Dev Results for Different Training Objectives}
    \label{tab:loss_func}
\end{table}

%% file: figs/politicians_guns.tex
\begin{tikzpicture}
\begin{axis}[
    width=1\textwidth,
    height=.7\textwidth,
    ybar stacked,
	bar width=15pt,
    legend style={
        style={font=\tiny},
        at={(0.5,-0.20)},
        anchor=north,legend columns=-1},
    symbolic x coords={Sanders, Biden, Trump},
    xtick=data,
    ymin=0,
    ymax=100,
    enlarge x limits=0.1,
    scaled y ticks=false,
    ymajorgrids=true,
    major x tick style=transparent,
    x tick label style={font=\scriptsize},
    y tick label style={font=\scriptsize}
    ]
\addplot+[ybar, style={one,fill=two,mark=none}] plot coordinates {(Sanders,68) (Biden,56) 
  (Trump,34)};
\addplot+[ybar, style={one,fill=five,mark=none}] plot coordinates {(Sanders,17) (Biden,18) 
  (Trump,13)};
\addplot+[ybar, style={one,fill=purple,mark=none}] plot coordinates {(Sanders,15) (Biden,26)
  (Trump,53)};
\legend{\strut Left, \strut Moderate, \strut Right}
\end{axis}
\end{tikzpicture}

%% file: figs/training_runtime.tex
\begin{tikzpicture}
    \begin{axis}[
        /pgf/number format/.cd,
        1000 sep={},
        width=1\textwidth,
        height=.65\textwidth,
        major x tick style=transparent,
        ybar=8pt,
        bar width=15pt,
        ymajorgrids=true,
        ylabel={Training in Seconds},
        label style={font=\Large},
        tick label style={font=\Large},
        symbolic x coords={AM, 4F, OD},
        xticklabels={Argument Mining, Stance Prediction, Open Domain},
        x tick label style={text width=10em, align=center},
        xtick=data,
        scaled y ticks=false,
        enlarge x limits=0.2,
        ymin=10,
        ymax=100000,
        ymode=log,
        point meta=rawy,
        log ticks with fixed point,
        log origin=infty,
        legend cell align=left,
        nodes near coords,
        every node near coord/.append style={font=\normalsize, color=black},
        legend columns=2,
        legend style={
                style={font=\Large},
                at={(0.27,0.75)},
                anchor=south west,
                /tikz/column 2/.style={column sep=3ex, font=\Large}
        }
    ]
    
        \addplot[style={one,fill=one,mark=none}]
            coordinates {(AM, 1153)(4F, 1078)(OD, 2615)};
    
        \addplot[style={purple,fill=purple,mark=none}]
            coordinates {(AM, 2277)(4F, 1029)(OD, 2260)};

        \addplot[style={six,fill=six,mark=none}]
            coordinates {(AM, 2220) (4F, 1241)(OD, 2262)};

        \addplot[style={two,fill=two,mark=none}]
             coordinates {(AM, 13293) (4F, 276) (OD, 65)};
             
        \addplot[style={five,fill=five,mark=none}]
             coordinates {(4F, 99) (OD, 326)};

        \legend{\Independent, DRaiL-ILP, DRaiL-AD$^3$, PSL, TensorLog}
    \end{axis}
\end{tikzpicture}

%% file: 9conclusion.tex
\section{Conclusions}
In this paper, we motivate the need for a declarative neural-symbolic approach that can be applied to NLP tasks involving long texts and contextualizing information. We introduce a general framework to support this, and demonstrate its flexibility by modeling problems with diverse relations and rich representations, and obtain models that are easy to interpret and expand. 
The code, data and documentation for \DRAIL and the application examples in this paper will be released to the community, to help promote this modeling approach for other applications.



%% file: 10acknowledgments.tex
\section*{Acknowledgments}

We would like to acknowledge current and former members of the PurdueNLP lab, particularly Xiao Zhang, Chang Li, Ibrahim Dalal, I-Ta Lee, Ayush Jain, Rajkumar Pujari and Shamik Roy for their help and insightful discussions in the early stages of this project. We also thank the reviewers and action editor for their constructive feedback. This project was partially funded by the NSF, grant CNS-1814105.

%% file: appendix.tex
\appendix

\section{DRaiL Programs}\label{appendix:programs}

\begin{figure}[H]
    \tiny
    
    \subfloat[a][\textbf{Issue-Specific Stance Prediction.}\\ \scriptsize{T: Thread, P: Post, I: Issue, A: Author}]{
    \begin{tabular}{l}
    \hline
        \textit{// Predictions} \\
        \textbf{r0:} \PRED{InThread(T,P)} $\Rightarrow$ \PRED{IsPro(P,I)}?\\
        
        \textbf{r1:} \PRED{Respond(P_2,P_1)} $\Rightarrow$ \PRED{Agree(P_1,P_2)}?\\ \\
         
        \textit{// Dependencies} \\
        \textbf{c0:} \PRED{Agree(P_1, P_2)?} $\wedge$ \PRED{IsPro(P_1,I)?} $\Rightarrow$ \PRED{IsPro(P_2,I)?}\\
        \textbf{c1:} \PRED{Agree(P_1, P_2)?} $\wedge$ \NPRED{IsPro(P_1,I)?} $\Rightarrow$ \NPRED{IsPro(P_2,I)?}\\
        \textbf{c2:} \NPRED{Agree(P_1, P_2)?} $\wedge$ \PRED{IsPro(P_1,I)?} $\Rightarrow$ \NPRED{IsPro(P_2,I)?}\\
        \textbf{c3:} \NPRED{Agree(P_1, P_2)?} $\wedge$ \NPRED{IsPro(P_1,I)?} $\Rightarrow$ \PRED{IsPro(P_2,I)?}\\
        \textit{// Author constraints} \\
        \textbf{c4:} \PRED{Author(P_1,A)} $\wedge$ \PRED{Author(P_2,A)} $\wedge$ \PRED{IsPro(P_1,I)?}
         $\Rightarrow$ \PRED{IsPro(P_2,I)?} \\
        \textbf{c4:} \PRED{Author(P_1,A)} $\wedge$ \PRED{Author(P_2,A)} $\wedge$ \NPRED{IsPro(P_1,I)?} 
         $\Rightarrow$ \NPRED{IsPro(P_2,I)?} \\
    
    \hline
    \end{tabular}}
    \\
    
    \subfloat[b][\textbf{Argument Mining.} \scriptsize{P: Paragraph, C: Component (Node), T: Type}]{\begin{tabular}{l}
    \hline
    \textit{// Predictions} \\
    \textbf{r0:} \PRED{InPar(C,P)} $\Rightarrow$ \PRED{NodeType(C,T)?} \\
    \textbf{r1:} \PRED{InPar(C_1,P)} $\wedge$ \PRED{InPar(C_2,P)} $\Rightarrow$ \PRED{Link(C_1,C_2)}? \\
    \textbf{r2:} \PRED{Link(C_1,C_2)}? $\Rightarrow$ \PRED{AttackStance(C_1,C_2)?}  \\
    \textbf{r3:} \PRED{InPar(C_1,P)} $\wedge$ \PRED{InPar(C_2,P)} $\wedge$ \PRED{InPar(C_3,P)} 
    $\Rightarrow$ \PRED{Grandp(C_1,C_2,C_3)?}  \\
    \textbf{r4:} \PRED{InPar(C_1,P)} $\wedge$ \PRED{InPar(C_2,P)} $\wedge$ \PRED{InPar(C_3,P)} 
    $\Rightarrow$ \PRED{Coparent(C_1,C_2,C_3)?}  \\
    \\
    \textit{// Higher order dependencies} \\
    \textbf{c0:} \PRED{Grandp(C_1,C_2,C_3)?} $\wedge$ \PRED{Link(C_1,C_2)?} $\Rightarrow$ \PRED{Link(C_2,C_3)?} \\
    \textbf{c1:} \PRED{Grandp(C_1,C_2,C_3)?} $\wedge$ \PRED{Link(C_2,C_3)?} $\Rightarrow$ \PRED{Link(C_1,C_2)?} \\
    \textbf{c2:} \PRED{Coparent(C_1,C_2,C_3)?} $\wedge$ \PRED{Link(C_1,C_3)?} $\Rightarrow$ \PRED{Link(C_2,C_3)?} \\
    \textbf{c3:} \PRED{Coparent(C_1,C_2,C_3)?} $\wedge$ \PRED{Link(C_2,C_3)?} $\Rightarrow$ \PRED{Link(C_1,C_3)?} \\
    \textit{// Source is always a premise} \\
    \textbf{c4:} \PRED{Link(C_1,C_2)?} $\Rightarrow$
    \PRED{NodeType(C_1, "Premise")?} \\
    \textit{// Multiclass constraint} \\
    \textbf{c5:} \PRED{HasType(C,+T)?} = 1 \\
    \textit{// Enforce tree structure} \\
    \textbf{c6:} \PRED{Link(C_1,+C_2)?} $\leq$ 1 \\
    \textbf{c7:} \PRED{Link(C_1,C_2)?} $\Rightarrow$ \PRED{Path(C_1,C_2)?} \\
    \textbf{c8:} \PRED{Path(C_1,C_2)?} $\wedge$ \PRED{Path(C_2,C_3)?} $\Rightarrow$ \PRED{Path(C_1,C_3)?} \\
    \textbf{c9:} \PRED{InPar(C_1,P)} $\wedge$ \PRED{InPar(C_2,P)} $\wedge$ \PRED{(C_1=C_2)} $\Rightarrow$ \NPRED{Path(C_1,C_2)?} \\
    \hline
    \end{tabular}
    }
   \caption{\DRAIL Programs}
\end{figure}

\section{Hyper-parameters}\label{appendix:hyperparams}

For BERT, we use the default parameters.

\begin{table}[H]
    \centering
    \resizebox{\columnwidth}{!}{%
    \begin{tabular}{llcc}
    \toprule
       Task & Param & Search Space & Selected Value \\
       \midrule
       Open Domain  & Learning Rate & 2e-6,5e-6,2e-5,5e-5 & 2e-5 \\
       (Local) & Batch size & 32 (Max. Mem) & 32 \\
       ~ & Patience & 1,3,5 & 3 \\
       ~ & Optimizer & SGD,Adam,AdamW &  AdamW \\
       ~ & Hidden Units & 128,512 & 512 \\
       ~ & Non-linearity & - & ReLU \\
       \midrule
       Open Domain & Learning Rate & 2e-6,5e-6,2e-5,5e-5 & 2e-6 \\
        (Global)  & Batch size & - & Full instance \\
       \midrule
       Stance Pred.  & Learning Rate & 2e-6,5e-6,2e-5,5e-5 & 5e-5 \\
       (Local) & Patience & 1,3,5 & 3 \\
        & Batch size & 16 (Max. Mem) & 16 \\
       ~ & Optimizer & SGD,Adam,AdamW &  AdamW \\
       \midrule
       Stance Pred.  & Learning Rate & 2e-6,5e-6,2e-5,5e-5 & 2e-6 \\
       (Global) & Batch size & - & Full instance \\
       \midrule
        Arg. Mining & Learning Rate & 1e-4,5e-4,5e-3,1e-3,5e-2,1e-2 & 5e-2 \\
        (Local)  & Patience & 5,10,20 & 20 \\
       ~ & Batch size & 16,32,64,128 & 64 \\
       ~ & Dropout & 0.01,0.05,0.1 & 0.05 \\
       ~ & Optimizer & SGD,Adam,AdamW &  SGD \\
       ~ & Hidden Units & 128,512 & 128 \\
       ~ & Non-linearity & - & ReLU \\
       \midrule
       Arg. Mining  & Learning Rate & 1e-4,5e-4,5e-3,1e-3,5e-2,1e-2 & 1e-4 \\
       (Global) & Patience & 5,10,20 & 10 \\
       ~ & Batch size & - & Full instance \\
       \bottomrule
    \end{tabular}}
    \caption{{\small Hyper-parameter Tuning \\ 
    }
}
    \label{tab:hyperparams}
\end{table}

\section{Code Snippets}\label{appendix:code}
\begin{figure}[H]
\centering
\subfloat[a][\DRAIL Program]{
    \includegraphics[width=0.5\textwidth]{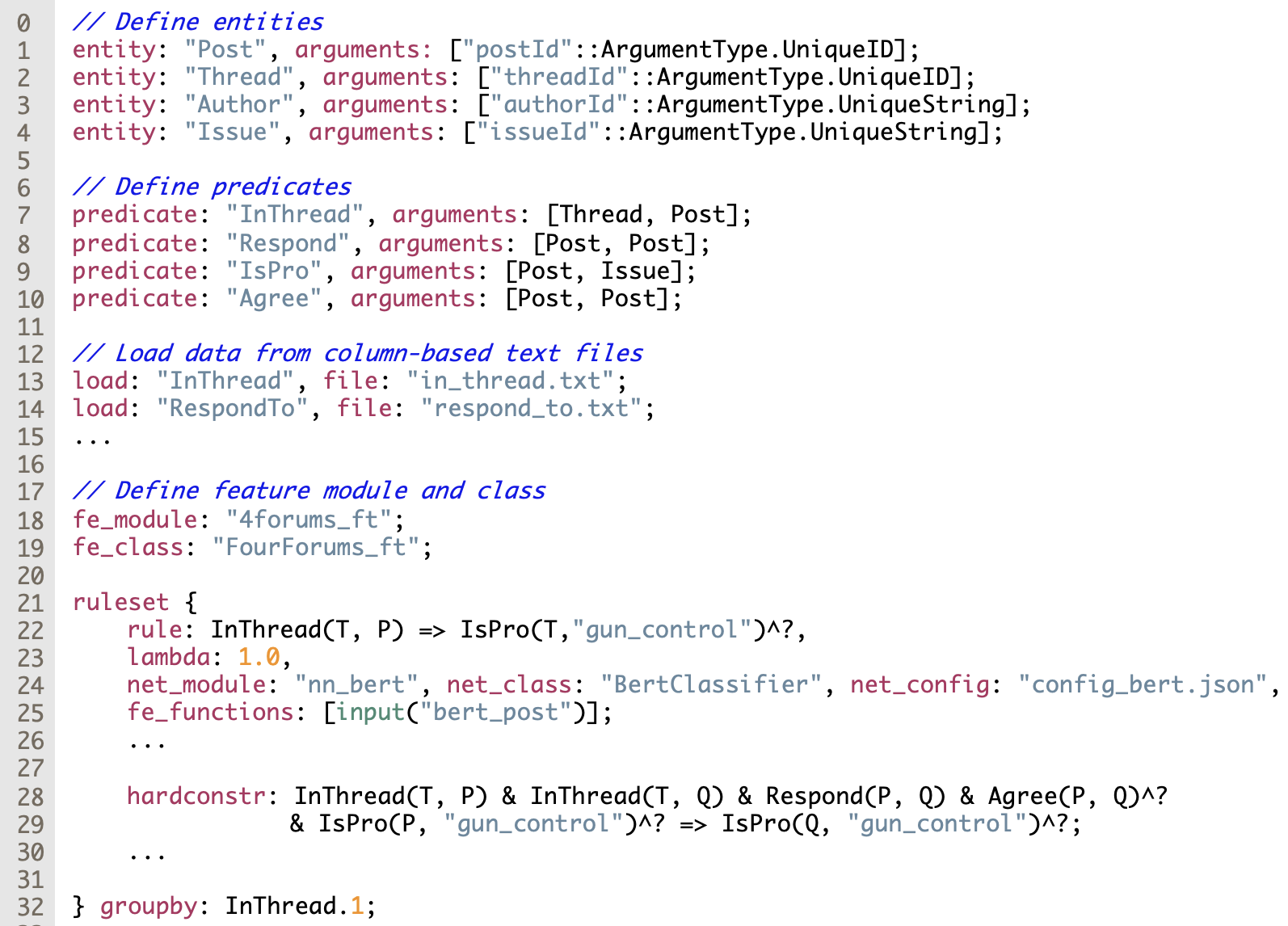}
}
\\

\subfloat[b][Neural Network Specification]{
    \includegraphics[width=0.5\textwidth]{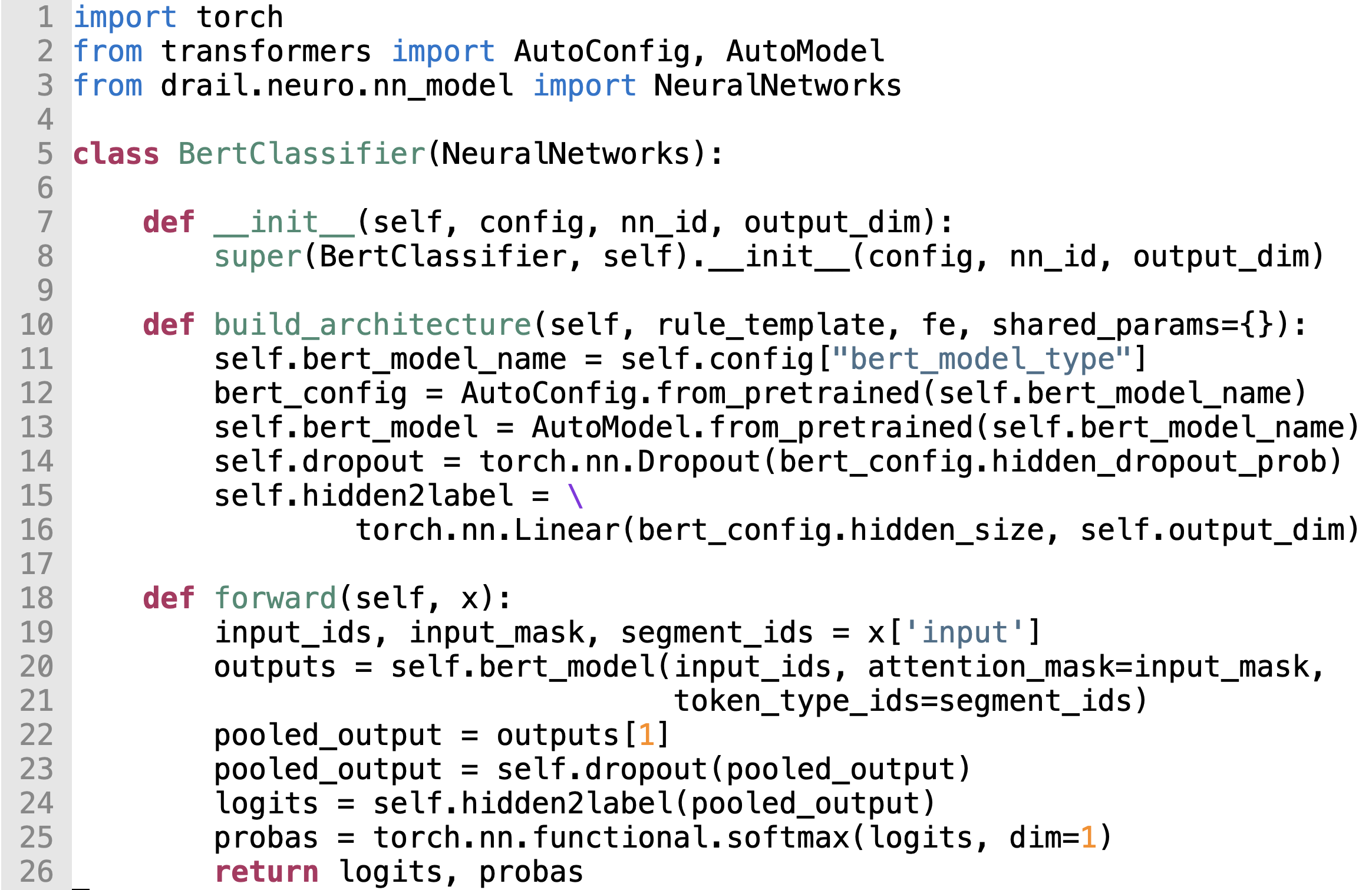}
}
\caption{Code Snippets}\label{fig:code_drail}
\end{figure}

We include code snippets to show how to load data into \DRAIL (Fig. \ref{fig:code_drail}-a), as well as to how to define a neural architecture (Fig. \ref{fig:code_drail}-b). 
Neural architectures and feature functions can be programmed by creating Python classes, and the module and classes can be directly specified in the \DRAIL program (lines 13, 14, 24 and 29 in Fig. \ref{fig:code_drail}-a).

